\documentclass[letterpaper]{article} % DO NOT CHANGE THIS
\usepackage{aaai24}  % DO NOT CHANGE THIS | [submission]
\usepackage{times}  % DO NOT CHANGE THIS
\usepackage{helvet}  % DO NOT CHANGE THIS
\usepackage{courier}  % DO NOT CHANGE THIS
\usepackage[hyphens]{url}  % DO NOT CHANGE THIS
\usepackage{graphicx} % DO NOT CHANGE THIS
\usepackage{natbib}  % DO NOT CHANGE THIS AND DO NOT ADD ANY OPTIONS TO IT
\usepackage{caption} % DO NOT CHANGE THIS AND DO NOT ADD ANY OPTIONS TO IT
\usepackage{amsmath,amssymb,amsthm}
\usepackage{algorithm}
\usepackage{algorithmic}
\usepackage{booktabs}
\usepackage{pifont}
\usepackage{multirow}
\usepackage{colortbl}
\usepackage[table,xcdraw]{xcolor}

\newtheorem{obs}{\textbf{Observation}}
\newcommand{\ours}{Graph Edit Distance Learning via Different Attention }
\newcommand{\ourshort}{REDRAFT}
\newcommand{\evalname}{Remaining Subgraph Alignment Test}
\newcommand{\evalshort}{RESAT}
\newcommand{\shared}{shared graph structure}
\newcommand{\unique}{graph structural difference}
\usepackage{newfloat}
\usepackage{listings}
\DeclareCaptionStyle{ruled}{labelfont=normalfont,labelsep=colon,strut=off} % DO NOT CHANGE THIS
\floatstyle{ruled}
\newfloat{listing}{tb}{lst}{}
\floatname{listing}{Listing}
\nocopyright
\title{Graph Edit Distance Learning via Different Attention}
\author{
    Jiaxi Lv$^{\rm 1}$   
    Liang Zhang$^{\rm 2}$   
    Yi Huang$^{\rm 1}$    
    Jiancheng Huang$^{\rm 1}$    
    Shifeng Chen$^{\rm 1}$\footnote{Corresponding author}\\
}
\affiliations{
    \textsuperscript{\rm 1}Shenzhen Institute of Advanced Technology\\
    \textsuperscript{\rm 2}Xidian University\\
    jx.lv1@siat.ac.cn
}

\usepackage{bibentry}
\begin{document}

\maketitle

\begin{abstract}
Recently, more and more research has focused on using Graph Neural Networks (GNN) to solve the Graph Similarity Computation problem (GSC), i.e., computing the Graph Edit Distance (GED) between two graphs. 
These methods treat GSC as an end-to-end learnable task, and the core of their architecture is the feature fusion modules to interact with the features of two graphs. Existing methods consider that \textbf{\textit{graph-level}} embedding is difficult to capture the differences in local small structures between two graphs, and thus perform fine-grained feature fusion on \textbf{\textit{node-level}} embedding to improve the accuracy, but leads to greater time and memory consumption in the training and inference phases. 
However, this paper proposes a novel \textbf{\textit{graph-level}} fusion module \textbf{Different Attention (DiffAtt)}, and demonstrates that graph-level fusion embeddings can substantially outperform these complex node-level fusion embeddings. We posit that the relative difference structure of the two graphs plays a key role in calculating their GED values. To this end, DiffAtt uses the difference between two graph-level embeddings as an attentional mechanism to capture the \unique{} of the two graphs. 
Based on DiffAtt, a new GSC method, named \textbf{G\underline{r}aph \underline{E}dit \underline{D}istance Lea\underline{r}ning vi\underline{a} Di\underline{f}ferent A\underline{t}tention (\ourshort{})}, is proposed, and experimental results demonstrate that \textbf{\ourshort{} achieves state-of-the-art performance in 23 out of 25 metrics in five benchmark datasets.} Especially on MSE, it respectively outperforms the second best by 19.9\%, 48.8\%, 29.1\%, 31.6\%, and 2.2\%. Moreover, we propose a quantitative test \textbf{\underline{Re}maining \underline{S}ubgraph \underline{A}lignment \underline{T}est (\evalshort{})} to verify that among all \textbf{\textit{graph-level}} fusion modules, the fusion embedding generated by DiffAtt can best capture the structural differences between two graphs. 
\end{abstract}

\section{Introduction}

Graph similarity computation (GSC) is a fundamental problem for graph-based applications, e.g., drug design \cite{tian2007saga}, graph similarity search \cite{zeng2009comparing,liang2017similarity}, and graph clustering \cite{wang2020inductive}. 
\textbf{Graph Edit Distance (GED)}, which is defined as the least number of graph edit operators to transform graph $G_i$ to graph $G_j$, is one of the most popular graph similarity metrics \cite{gao2010survey,neuhaus2006fast,bougleux2015quadratic}. 
Unfortunately, the exact GED computation is NP-Hard in general \cite{zeng2009comparing}, which is too expensive to leverage in practice. 

Recently, many Graph Neural Networks (GNNs) based GSC methods have been proposed to compute the GED in a faster manner \cite{bai2019simgnn,bai2020learning,li2019graph,ling2021multilevel,qin2021slow,zhang2021h2mn,bai2021tagsim,wang2021combinatorial,ranjan2022greed,zhuo2022efficient}. 
The GNN-based algorithms transform the GED value to a similarity score and use an end-to-end framework to learn to map the given two graphs to their similarity score. 
As a general framework, the Siamese neural network is used to aggregate information on each graph, while the feature fusion module is used to capture the graphs' similarity representation, and then the Multi-layer Perceptron (MLP) is leveraged for the similarity score regression. 

As discussed by \citeauthor{qin2021slow}, capturing the interaction between two graphs is instrumental in tackling the GSC problem. Thoroughly modeling the intricate interplay between graphs is crucial for accurately characterizing their similarity. 
Most of the existing GSC methods drop the graph-level fusion module and instead interact on node-level embeddings to learn the similarities between two graphs in a more fine-grained way, i.e., computing the similarity or attention between all nodes of two graphs \cite{li2019graph,bai2020learning,ling2021multilevel}. 
It is believed that the differences between one graph and other graphs may be on any small local structure, but the graph-level embedding of each graph is a fixed one vector and it is difficult to reflect these small differences information flexibly \cite{bai2020learning}. 
\textit{However, node-level fusion entails computational and memory costs that scale quadratically with the number of nodes, presenting challenges for large-scale graphs.} 
To address these issues, the Alignment Regularization (AReg) \cite{zhuo2022efficient} imposes constraints on the GNN Encoder during training to enable it to learn representations that capture the underlying alignment between nodes across a pair of graphs. This helps the model learn cross-graph interactions in graph-level embeddings without needing to explicitly compute node-to-node similarities. 
\textit{However, the regularization strength hyperparameter $\lambda$ of AReg needs proper tuning to prevent the regularization term from adversely affecting the model.}

Developing graph-level fusion modules tailored for GSC still remains an open challenge. Many existing designs like Neural Tensor Networks (NTN) \cite{socher2013reasoning} or Embedding Fusion Network (EFN) \cite{qin2021slow} are largely based on intuition rather than principled analysis. 
However, we find in Sec \ref{section:anaysis_on_graph_fusion} that \textit{the GED is only correlated with the discrepancy in the two graph structures and thus concludes that the graph-level fusion module should capture the structural differences between two graphs.} 
Inspired by this, we propose a novel graph-level fusion module \textbf{DiffAtt}. Especially, DiffAtt generates an attention vector reflecting the relative structural differences between two graphs, and this attention is leveraged to update each graph-level representation via element-wise multiplication, resulting in new representations encoding the structural differences. 
Moreover, we propose a novel GSC architecture that outperforms existing node-level fusion models and achieves state-of-the-art performance. To sum up, the contributions of this paper can be summarized as follows: 
\begin{enumerate}

\item A novel graph-level fusion module, \textbf{DiffAtt} is proposed, which is an attention mechanism to highlight structural differences between two graphs. Experimental results demonstrate that DiffAtt has strong generalization and can substantially improve the performance of the model. 
\item Based on DiffAtt, a GSC model, \textbf{\ours{}(\ourshort{})} is proposed. In addition to the DiffAtt, \ourshort{} uses multi-scale GIN layers to extract more representative graph-level embeddings, each of which contains residual connections and an additional Forward Forward Network (FFN) to enhance node features. 
Compared with current methods on five benchmark datasets, \ourshort{} achieves state-of-the-art performance on 23 out of 25 metrics. 
\item To verify the ability to extract \unique{} information, a quantitative test \textbf{\evalname{} (\evalshort{})} is proposed, and the results demonstrate that the ability of the graph-level fusion module to extract \unique{} information is positively correlated with the performance on GSC, while \textbf{DiffAtt} has the strongest ability to extract \unique{} information.

\end{enumerate}

\section{Preliminary and Related Work}
\label{related_work}
In this section, we introduce the notations, the definitions of GED and GSC, and the related works.

\textbf{Notations.} The graph data $G$ can be viewed as a pair of the set of nodes $V = \{v_i\}_{i = 1}^N$ and the set of edges $E$. The $|V| = N$ represents the number of nodes. In our setting, all graphs are undirected and edges have no attributes. Therefore, the structure of the graph can be represented using the adjacency matrix $A \in \{0,1\}^{N \times N}$ and the nodes $v_i$ and $v_j$ are connected by an edge if and only if $A_{i,j} = 1$. The node feature matrix is $X \in \mathbb{R}^{N \times C}$. $C$ is the dimension of its features. $x_i = \xi(v_i)$, where $\xi(\cdot)$ maps node $v_i$ to node feature $x_i \in \mathbb{R}^{C}$.

\textbf{Graph Edit Distance (GED).} The graph edit distance (GED) is a measure of the minimum \textit{\textbf{dissimilarity}} between two graphs $G_1$ and $G_2$, which is defined as the minimum cost of transforming $G_1$ to become isomorphic with $G_2$. Specifically, a graph edit operator $e_i$ can be a node or edge insertion, removal, or substitution. An edit path $\mathcal{P}(G_1,G_2)$ is a sequence of graph edit operators that transform $G_1$ into a graph isomorphic to $G_2$. The graph edit distance (GED) is defined as:
\begin{equation}
\label{ged_defintion}
GED(G_1, G_2) = \min_{\{e_1, e_2, ..., e_k\} \in \mathcal{P}(G_1,G_2)} \sum_{i=1}^k c(e_i)
\end{equation}
where $c(e_i)$ denotes the real non-negative cost function of the $i$-th edit operator $e_i$ in the edit sequence $\{e_1, e_2, ..., e_k\}$. 
We follow the setting of prior work \cite{bai2019simgnn}, defining all $c_e(\cdot)$ as 1.

\textbf{Graph Similarity Computation Problem (GSC).} 
GED can effectively capture the edge and node differences between two graphs and has been widely used in real-world scenarios \cite{bai2019simgnn,zhuo2022efficient}. Therefore, the GSC adopts GED as the ground truth similarity value. 
In order to better conform to the learning paradigm of GNNs, the GED is normalized as $nGED(G_i,G_j) = \frac{GED(G_i,G_j)}{(|N_i|+ |N_j|)/2}$, and the similarity score between two graphs is defined as $s(G_i,G_j) = exp(-nGED(G_i,G_j))$, which is in the range of (0,1]. 
The GSC is defined as: given two graphs $G_i$ and $G_j$ with their similarity metric, the GSC models learn a function that maps the two graphs to their similarity score $s(G_i,G_j)$.

\textbf{Related Work.} 
The Graph Matching Network (GMN) proposed by \citeauthor{li2019graph} is the first GNN-based GSC model. It computes similarity via cross-graph node-to-node attention mechanisms. 
\citeauthor{bai2019simgnn} proposed the SimGNN model, which uses GCN layers \cite{kipf2016semi} to update node embeddings and uses the Neural Tensor Networks (NTN) \cite{socher2013reasoning} module to learn graph fusion embeddings in the two graph-level embeddings. 
However, their later work \cite{bai2020learning} argued that the small local structural differences are difficult to be captured by a single graph embedding vector, and thus proposed GraphSim, which directly learns the similarity based on the node-level interaction of the two given graphs. 
Motivated by this, H2MN \cite{zhang2021h2mn} designed a hierarchical node-level interaction module, while MGMN \cite{ling2021multilevel} proposed a more complex node-graph level interaction module. 
Many recent works have focused on designing complex, fine-grained fusion modules, under the belief that modeling node-level interactions is key for accuracy. 
However, relying on the complex node-level fusion entails significant costs in memory, training time, and inference speed due to the quadratic growth in computational complexity. 
To address this, \citeauthor{zhuo2022efficient} proposed the Alignment Regularization technique that enables graph-level embeddings to better capture cross-graph interactions without needing expensive node-level fusion. By regularizing the GNN Encoder during training, their method aligns the graph embeddings to remove the need for direct node-level feature comparisons, demonstrating that this graph-level approach can achieve competitive or better performance compared to node-level fusion. 

Our work presented the same conclusion as \citeauthor{zhuo2022efficient}, showing that feature fusion on graph-level embeddings can also lead node-level or node-graph-level fusion embeddings substantially. 
However, the difference of \ourshort{} from the previous approaches is: 1). \ourshort{} uses a new graph-level fusion module DiffAtt to better explore the differences between the two graph structures; 2). \ourshort{} uses \textit{residual connections} to help optimize the model training process, and adds the \textit{FFN} to augment the nonlinear transformations to learn a more discriminative representation of the graph-level embeddings.

\section{Motivation}
\label{section:anaysis_on_graph_fusion}

\begin{figure}[t]
% %\vspace{-3.0em}
  \centerline{\includegraphics[width=0.5\textwidth]{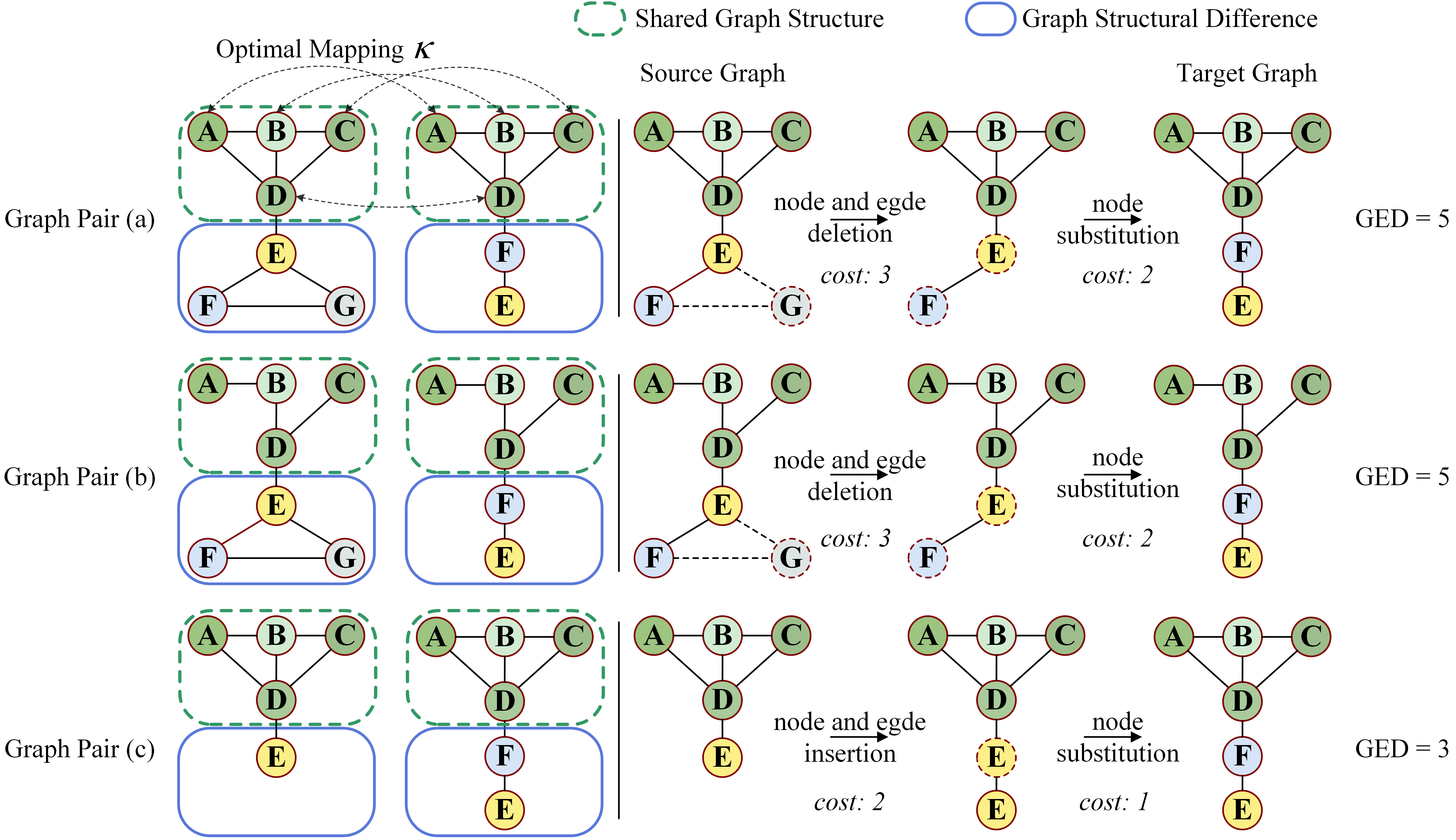}}
  %\vspace{-0.5em}
  \caption{Graph edit processes for some graph pairs.}
  \label{pic_motivation}
\end{figure}

Compared to other graph learning tasks, the GSC model needs to effectively fuse the features of two graphs to learn their similarity. However, there has been little principled analysis on what role a graph fusion module should fulfill to enable accurate GED computation. In this section, we analyze the GED and graph matching problem, derive the key responsibilities of the graph-level fusion module, and discuss the existing graph-level fusion modules. 

GED is a form of graph matching problem. Graph matching aims to find an optimal mapping that maximizes the similarity between two graphs. 
However, GED as a measure of the minimum difference between two graphs, focuses more on the structural difference between two graphs under their optimal mapping. 
Assume that the optimal bijective mapping between two graphs $G_1$ and $G_2$ is $\kappa: \hat{V}_1 \rightarrow \hat{V}_2$ between $\hat{V}_1 \subseteq V_1$ and $\hat{V}_2 \subseteq V_2$ such that node features are preserved, i.e., $\xi(v) = \xi(\kappa(v))$. 
Specifically, we define two edge sets $\hat{E}_1 \subseteq E_1$ and $\hat{E}_2 \subseteq E_2$ that satisfy if $(u,v) \in \hat{E}_1$, where $u,v \in \hat{V}_1$, these must exist $(\kappa(u),\kappa(v)) \in \hat{E}_2$. 
The optimal mapping between two graphs divides the two graphs into two separate components, respectively:
\begin{enumerate}
    \item The \textbf{\textit{\shared{}}} component, represented by $(\hat{V}_1,\hat{E}_1)$, containing the substructures that is common between both graphs.
    \item The \textbf{\textit{\unique{}}} component, e.g., $(V_1 - \hat{V}_1,E_1 - \hat{E}_1)$ for $G_1$, containing substructure that one graph holds and the other does not. 
\end{enumerate}

We observe that graph editing operations are not performed on the \shared{}. Therefore, we can derive a key insight into the role of \shared{} versus \unique{} in determining GED values: 
\begin{obs}
\label{section:obs_ged}
Under the optimal mapping, the GED is solely determined by the \unique{} of the two graphs, and their \shared{} have no impact on their GED value. 
\end{obs}

Fig \ref{pic_motivation} shows the impact of \shared{} and \unique{} between several graph pairs on the GED values. Although graph pairs (a) and (b) have different \shared{} parts, \textit{they have the same \unique{} parts and thus the same optimal edit path and GED value.} 
In contrast, graph pairs (a) and (c) have the same \shared{} parts, \textit{but different \unique{} parts, leading to different optimal edit paths and GED values.} \textit{Based on this insight, we hypothesize that the fusion module can enable more accurate GED modeling by focusing on capturing \unique{} rather than \shared{}.} 

However, most graph-level fusion modules do not directly aim to extract \unique{}. 
A widely used learnable graph-level fusion module is the Neural Tensor Networks (NTN) module \cite{socher2013reasoning,bai2019simgnn,zhuo2022efficient}, which is defined as:
\begin{equation}
    NTN(\boldsymbol{h}_i,\boldsymbol{h}_j) = f(\boldsymbol{h}_i^T W^{[1:K]} \boldsymbol{h}_j + V [\boldsymbol{h}_i,\boldsymbol{h}_j] + b ),
\end{equation}
where $\boldsymbol{h}_i$ and $\boldsymbol{h}_j$ denote the embeddings of graph $G_i$ and $G_j$, respectively. $W^{[1:K]} \in \mathbb{R}^{C \times C \times K}$, $V \in \mathbb{R}^{K \times 2C}$ and $b \in \mathbb{R}^{K}$ are the learnable parameters. The advantage of the NTN module lies in allowing the interaction between two graph-level features through multiplication in the first term \cite{socher2013reasoning}. However, if the learnable parameter $W^{[1:K]}$ is assumed to be an identity matrix, the first term of NTN can degenerate into the inner product of two graph embeddings, which indicates that \textit{\textbf{NTN may prefer to compute similarities rather than differences between two graphs.}} 
% \textbf{\textit{NTN focuses more on \shared{} rather than \unique{}}}. 
The Embedding Fusion Network (EFN) is another graph-level fusion module, which is defined as $EFN(\boldsymbol{h}_{ij}) = MLP(\sigma(W_U \delta(W_D \boldsymbol{h}_{ij}))\cdot \boldsymbol{h}_{ij} + \boldsymbol{h}_{ij})$, 
where $\boldsymbol{h}_{ij}$ denotes the concatenation of the two graph embeddings. $\sigma$ and $\delta$ denote the sigmoid gating and ReLU function, respectively. 
Although EFN enhances the fusion representation of two graph embeddings through MLPs, \textbf{\textit{it does not selectively highlight \unique{} features}}. 
In addition to NTN and EFN, the element-wise absolute distance or squared distance is another commonly used fusion module, which is defined as $|\boldsymbol{h}_i - \boldsymbol{h}_j|^{p}$, where the order $p$ is respectively set to 1 or 2. 
While these methods aim to highlight \unique{} by directly computing the difference between the embeddings, \textbf{\textit{they perform poorly in practice, likely because subtractive operations discard the rich details within \unique{}}}. 

However, this paper proposes a novel graph-level fusion module, named \textbf{Different Attention (DiffAtt)}. 
DiffAtt leverages the difference between the two embeddings as a form of attention to highlight \unique{} features in graph-level embeddings, which allows for more \unique{} detailed features to be retained. 
We find that DiffAtt can substantially improve performance, outperforming even complex node-level fusion models. 
Furthermore, as validated quantitatively in Sec \ref{remaining_subgraph_test}, DiffAtt appears to capture detail \unique{} information better than other graph-level fusion methods and therefore can better improve the performance of the model. 

\begin{figure*}[t]
  \centerline{\includegraphics[width=0.8\textwidth]{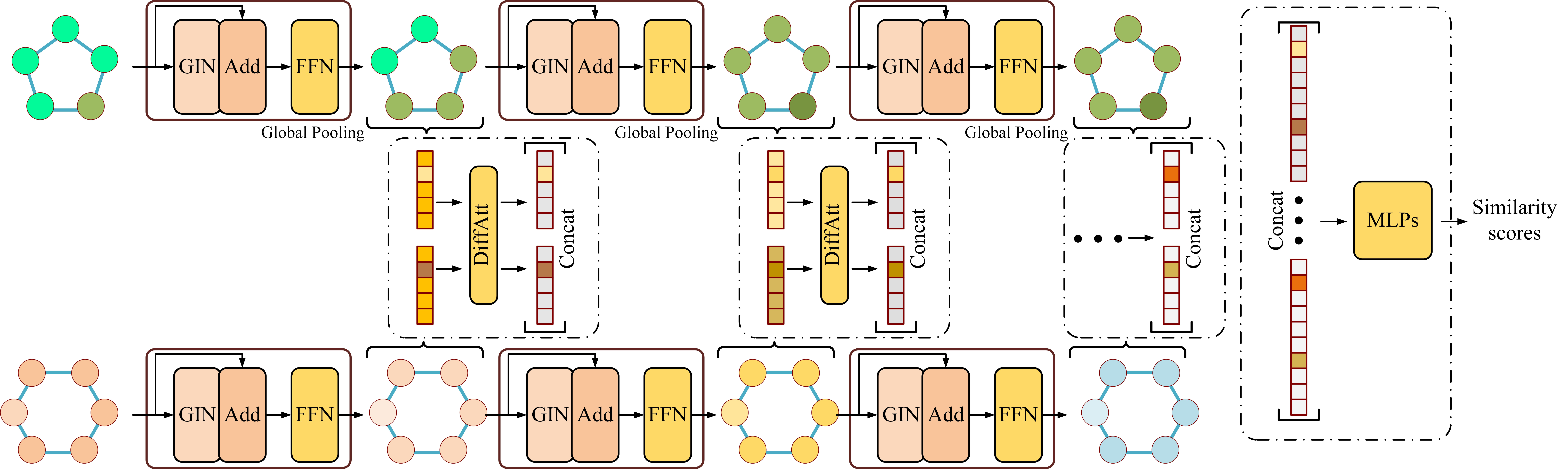}}
  \caption{\ourshort{} first uses the multi-scale GIN Encoder to aggregate the information in the graph, then DiffAtt for feature fusion, and finally MLP to predict the similarity scores.}
  \label{n2agim_model_pic}
\end{figure*}

\section{Proposed Methods}
\label{proposed_methods}

In this section, we first introduce the \textbf{\textit{DiffAtt}} and then present the overall architecture of the \textit{\textbf{G\underline{r}aph \underline{E}dit \underline{D}istance Lea\underline{r}ning vi\underline{a} Di\underline{f}ferent A\underline{t}tention (\ourshort{})}} model.

\subsection{Different Attention (DiffAtt)} 
As motivated in Sec \ref{section:anaysis_on_graph_fusion}, DiffAtt aims to highlight the relatively \unique{} of the two graphs. 
Although directly using the distance between two embeddings can encode differences, it may lose fine-grained details about graph structural differences. Since rich graph structural difference features are already contained in the graph-level embeddings, DiffAtt uses the discrepancy between the graph-level embeddings to guide an attention mechanism that amplifies \unique{} features, while suppressing \shared{} features. 

\textit{First}, we take the two graph embeddings $\boldsymbol{h}_i$ and $\boldsymbol{h}_j$ and calculate the element-wise absolute difference between them. This gives us a new representation that encodes the variation between the graph embeddings. \textit{Second}, we pass this through a MLP to adjust and refine the difference information as $\boldsymbol{h}_{diff} = MLP(|\boldsymbol{h}_i - \boldsymbol{h}_j|)$. 
\textit{Next}, a softmax function converts these differences into attention weights as $\boldsymbol{\alpha} = softmax\left(  \boldsymbol{h}_{diff} \cdot t^{-1}  \right)$, with higher values indicating dimensions that diverge more across the graphs, and $t$ is the temperature factor to control the smoothness of the difference. 
\textit{Finally}, we multiply $\boldsymbol{\alpha}$ element-wise with the original graph embeddings as $\boldsymbol{u}_{G_i} = \boldsymbol{\alpha} \odot \boldsymbol{h}_i $ and $\boldsymbol{u}_{G_j} = \boldsymbol{\alpha} \odot \boldsymbol{h}_j$. This selectively amplifies the dimensions of \unique{}, while suppressing aspects of \shared{}. 

\subsection{Architecture of \ourshort{}}

Based on the DiffAtt, we propose a new GSC model named \ours{} (\ourshort{}). 
The architecture of \ourshort{} is $f(G_i,G_j) = \Theta(F(E(G_1),E(G_2)))$, where $E(\cdot)$ is a GNN Encoder that generates more representative embeddings for each graph independently, $F(\cdot,\cdot)$ is a graph-level fusion module, implemented as our proposed DiffAtt, and $\Theta(\cdot)$ is an MLP regressor that maps the fused embedding generated by $F$ to the similarity score. 
Fig \ref{n2agim_model_pic} shows the block diagram of the \ourshort{} and the details are as follows:

In order to better leverage DiffAtt, we believe that our GNN Encoder should satisfy two properties. The first one is the need to satisfy the \textbf{\textit{permutation invariance}} to implicitly align two graphs. 
Finding the optimal mapping between two graphs is equivalent to permuting nodes to match across graphs. By making the GNN Encoder permutation invariant, it can implicitly align the two input graphs when encoding them using shared parameters. This allows DiffAtt to better focus on modeling the \unique{} between the aligned graphs. 
The second one is having \textbf{\textit{strong expressive power}} to extract graph detail information in graph-level embeddings, which can help DiffAtt better capture the information of \unique{}.

The Graph Isomorphism Network (GIN) \cite{xu2018powerful} has the same expressive power as the Weisfeiler-Lehman (WL) graph isomorphism test, i.e. determining whether two graphs are non-isomorphic. 
Therefore, we choose GIN as our GNN layer to update the embedding of the nodes, which is defined as: 
\begin{equation}
GIN(x_{i}^{(k)}) = MLPs \left( \left(1+\epsilon \right) \cdot x_{i}^{(k)} + \sum_{j \in \mathcal{N}(i)} x_{j}^{(k)} \right), 
\end{equation}
where $\epsilon$ is a learnable parameter or a fixed scalar and $k$ stands for at the k$^{th}$ layer. 
$\mathcal{N}(i)$ denotes the neighbors of node $i$. 
All the MLPs in GIN have one linear layer with the Layer Normalization \cite{ba2016layer} and ReLU activation function. 
\textit{To promote model training optimization and boost the expressiveness of node embeddings, we enhance GIN layers with Residual Connections \cite{he2016deep} and FeedForward Neural Network (FFN) as} $x_{i}^{(k+1)} = FFN(x_{i}^{(k)} + GIN(x_{i}^{(k)}))$. 
Lower layers capture local neighborhood information while higher layers capture more global patterns. Thus, we use $K$ layers of GIN to aggregate information from neighboring nodes in multiple stages, and extract the features $X^{k}$ from the output of each layer of GIN. 
The learnable Global Context-Aware Attention (GCA) \cite{bai2019simgnn} is leveraged as our graph-level readout function, because it learns the attention of nodes based on global context information to obtain graph-level representations with strong expressive power and discriminativity. GCA is defined as:
\begin{equation}
\begin{aligned}
& \boldsymbol{c}^{(k)} = tanh((\frac{1}{N}\sum^{N}_{i=1} x_i^{(k)} )W^{(k)}) \\
& \boldsymbol{h}^{(k)} = \mathcal{F}_{CBA}(X^{(k)}) = \sum_{i = 1}^{N} \sigma({\boldsymbol{x}^{(k)}_{i}}^{T} \boldsymbol{c}^{(k)}) \boldsymbol{x}_i^{(k)},
\end{aligned}
\end{equation}
where $\boldsymbol{c}^{(k)}$ denotes the context information of the graphs, and $W^{(k)} \in \mathbb{R}^{C^{(k)} \times C^{(k)}}$ is the learnable parameter. \textit{Based on this architecture, our GNN Encoder not only satisfies permutation invariance but also generates more expressive graph-level embeddings.} 

DiffAtt is applied to the graph-level embeddings obtained at each layer to generate fused embeddings at multi-scale, and all the enhanced embeddings $\boldsymbol{u}_{G_i}^{(k)}$ and $\boldsymbol{u}_{G_j}^{(k)}$ are concatenated to obtain the multi-scale fused embedding as $\boldsymbol{u}_{G_i,G_j} = concat([\boldsymbol{u}_{G_i}^{(0)},\boldsymbol{u}_{G_j}^{(0)}, \cdots, \boldsymbol{u}_{G_i}^{(K)},\boldsymbol{u}_{G_j}^{(K)}])$. A MLP with two hidden layers and ReLU activation is then applied to map $\boldsymbol{u}_{G_i,G_j}$ to the similarity score. 
We do not use a sigmoid output to avoid training instability, and we adopt the \textbf{Mean Square Error (MSE)} as the loss function to train \ourshort{}.

\textbf{Time Complexity Analysis.} The time complexity of GIN is $\mathcal{O}(max(E_i,E_j))$ and the time complexity of the FFN, DiffAtt, and MLP regressors are all $\mathcal{O}(C \times C)$. Compared to cross-graph node-level interactions with complexity $\mathcal{O}(N_i \times N_j)$, \ourshort{}'s DiffAtt has lower complexity and faster inference, especially on large-scale graphs.

\section{Experiments}
\label{exp}
In this section, we evaluate our method using the five benchmark datasets provided by \citet{bai2019simgnn,bai2019unsupervised} in GSC and compare our method with other methods. 

The \textbf{\ourshort{}} is evaluated using the PyTorch Geometric \cite{fey2019fast}. 
Following the setup of the previous work, we use the Adam optimizer with a learning rate of 0.001 and batch size set to 128. All hidden channels are set to 64 in \ourshort{}. Typically, we run 18 epochs on each dataset and perform 20 validations uniformly on the last epochs. Since the NCI109 dataset is relatively large, we only run 10 epochs. 
Ultimately, the parameter that results in the least validation loss is chosen to perform the evaluations on the test data. 
Note that all the experiments are performed on a Linux server with Intel(R) Xeon(R) CPU E5-2690 v4 @ 2.60GHz and 8 NVIDIA GeForce RTX 3090Ti. 
The experiments are conducted on five widely used graph similarity benchmark datasets: AIDS700nef \cite{bai2019simgnn}, LINUX \cite{wang2012efficient}, IMDBMulti \cite{yanardag2015deep}, PTC \cite{shrivastava2014new} and NCI109 \cite{wale2008comparison}. We evaluate model performance using several common metrics: Mean Square Error (MSE, in the format of $10^{-3}$), Spearman's Rank Correlation Coefficient ($\rho$) \cite{spearman1961proof}, Kendall's Rank Correlation Coefficient ($\tau$) \cite{kendall1938new}, and Precision at $k$ (P@$k$) \cite{bai2019simgnn}. Details of the datasets, data processing, evaluation metrics, and other details are included in the Appendix. All code and trained model parameters are available in the supplementary material.

\begin{table}[t]

    \centering
    \resizebox{0.5\textwidth}{1.4cm}{
    \begin{tabular}{cccccccccccccc}
    \toprule
         \multirow{2}*{Readout} & \multirow{2}*{DiffAtt}  & \multicolumn{2}{c}{AIDS700nef} & \multicolumn{2}{c}{LINUX} & \multicolumn{2}{c}{IMDBMulti} & \multicolumn{2}{c}{PTC} & \multicolumn{2}{c}{NCI109}   \\
    % \midrule
         &  & MSE $\downarrow$  & $\rho$ $\uparrow$ & MSE $\downarrow$ & $\rho$ $\uparrow$ & MSE $\downarrow$ & $\rho$ $\uparrow$ & MSE $\downarrow$ & $\rho$ $\uparrow$ & MSE $\downarrow$ & $\rho$ $\uparrow$  \\
    % \midrule
    \cmidrule(r){3-4}  \cmidrule(r){5-6} \cmidrule(r){7-8}  \cmidrule(r){9-10} \cmidrule(r){11-12}
        \multirow{2}*{\textit{Mean}}& \ding{56}& 3.027  & 0.812 & 0.541 & 0.985 & 0.495 & 0.864 & 5.707 & 0.746 & 4.596 & 0.516  \\
        % ~ & \ding{52}& 2.075  & 0.864& 0.311& 0.989& 0.343& 0.928& 4.711& 0.791& 4.388& 0.540 \\
        ~  & \ding{52} & 2.075$^\dagger$ & 0.864$^\dagger$ & 0.311$^\dagger$ & 0.989$^\dagger$ & 0.343$^\dagger$ & 0.928$^\dagger$ & 4.711$^\dagger$ & 0.791$^\dagger$ & 4.388$^\dagger$ & 0.540$^\dagger$ \\

        \midrule
        \multirow{2}*{\textit{Max}}& \ding{56}& 3.225 & 0.805 & 0.356 & 0.989 & 0.403 & 0.853 & 4.385$^\dagger$& 0.818 & 4.030 & 0.595$^\dagger$ \\
        % & \ding{52}& 2.060  & 0.855& 0.232& 0.991& 0.311& 0.920& 4.414 & 0.821& 3.922& 0.584 \\
         & \ding{52} & 2.060$^\dagger$ & 0.855$^\dagger$ & 0.232$^\dagger$ & 0.991$^\dagger$ & 0.311$^\dagger$ & 0.920$^\dagger$ & 4.414 & 0.821$^\dagger$ & 3.922$^\dagger$ & 0.584 \\

        \midrule
        \multirow{2}*{\textit{Sum}}& \ding{56}& 1.427 & 0.901 & 0.122 & 0.992 & 0.387 & 0.859 & 1.539 & 0.944 & 3.650 & 0.680 \\
        % & \ding{52}& 1.096& 0.921& 0.062& \textbf{0.994}& 0.316& \textbf{0.929}& 1.040& 0.961& \textbf{3.587}& \textbf{0.686} \\
         & \ding{52} & 1.096$^\dagger$ & 0.921$^\dagger$ & 0.062$^\dagger$ & \textbf{0.994}$^\dagger$ & 0.316$^\dagger$ & \textbf{0.929}$^\dagger$ & 1.040$^\dagger$ & 0.961$^\dagger$ & \textbf{3.587}$^\dagger$ & \textbf{0.686}$^\dagger$ \\

        \midrule
        \multirow{2}*{\textit{GCA}}& \ding{56}& 1.389 & 0.906 & 0.115 & 0.992 & 0.377 & 0.856 & 1.655 & 0.938 & 3.501$^\dagger$& 0.671  \\
        % & \ding{52}& \textbf{1.076}& \textbf{0.923}& \textbf{0.048}& \textbf{0.994}& \textbf{0.296}&0.914& \textbf{1.015}& \textbf{0.962}& 3.727& 0.675 \\
         & \ding{52} & \textbf{1.076}$^\dagger$ & \textbf{0.923}$^\dagger$ & \textbf{0.048}$^\dagger$ & \textbf{0.994}$^\dagger$ & \textbf{0.296}$^\dagger$ &0.914$^\dagger$ & \textbf{1.015}$^\dagger$ & \textbf{0.962}$^\dagger$ & 3.727 & 0.675$^\dagger$ \\
    \bottomrule
        \end{tabular}}
\caption{Experimental results on the efficiency and generalization of DiffAtt. The $\dagger$ represents the best performance with or without DiffAtt, while the bold denotes the best performance of the four graph readout functions. The $\uparrow$ means that the larger this indicator is, the better the performance, while the $\downarrow$ indicates the opposite.}
\label{exp:diffatt_with_diff_pooling}
\end{table}

\begin{table}[t]

    \centering
    \resizebox{0.45\textwidth}{1.5cm}{
    
    \begin{tabular}{cccccccccc}
    \toprule
         \multirow{2}*{GNN} & \multirow{2}*{$t$}  & \multirow{2}*{ENH} & \multicolumn{2}{c}{AIDS700nef} & \multicolumn{2}{c}{IMDBMulti} & \multicolumn{2}{c}{PTC} \\
          ~ & ~  & ~  &  MSE $\downarrow$  & $\rho$ $\uparrow$ &  MSE $\downarrow$  & $\rho$ $\uparrow$ & MSE $\downarrow$  & $\rho$ $\uparrow$  \\ 
        \cmidrule(r){4-5}  \cmidrule(r){6-7} \cmidrule(r){8-9}
    GCN & 1 & \ding{52}  & 1.077 & 0.921 & 0.313 & 0.938$^\dagger$ & 1.059 & 0.961  \\
    GAT & 1 & \ding{52}  & 1.147 & 0.913 & 0.332 & 0.901 & 1.029 & \textbf{0.962}$^\dagger$  \\
    % \midrule
    GIN & 1 & \ding{56}  & 1.238 & 0.909 & 0.717 & 0.930 & 1.054 & 0.960  \\
    GIN & 1 & \ding{52}  & 1.076$^\dagger$ & 0.923$^\dagger$ & \textbf{0.296}$^\dagger$ & 0.914 & 1.015$^\dagger$ & \textbf{0.962}$^\dagger$ \\

    \midrule
    
    GIN & 0.5 & \ding{52} & 1.159 & 0.919 & 0.303 & \textbf{0.958} & 1.056 & \textbf{0.962} \\
    
    GIN & 2 & \ding{52}  & 1.052 & 0.923 & 0.319 & 0.951 & 1.029 & 0.955 \\
    
    GIN & 4 & \ding{52}  & 1.051 & 0.923 & 0.310 & 0.940 & 0.995 & 0.951 \\
    
    GIN & \textit{learnable} & \ding{52}  & \textbf{1.037} & \textbf{0.925} & 0.299 & 0.904 & \textbf{0.993} & \textbf{0.962} \\ 
    \bottomrule
    \end{tabular}
}
\caption{Experimental results on using different GNNs, temperature factor $t$, and enhancement (ENH). $\dagger$ represents the best performance in the first four experiments. The bold denotes the best performance with all configurations.}
\label{exp:gnn_enhance_learnable}
\end{table}

\begin{table*}[t]

\resizebox{1.0\textwidth}{1.15cm}{
\begin{tabular}{cccccccccccccccccccccccccc}
\toprule
         & \multicolumn{5}{c}{AIDS700nef}                                                                                                                  & \multicolumn{5}{c}{LINUX}                                                                                               & \multicolumn{5}{c}{IMDBMulti}                                                                                                       & \multicolumn{5}{c}{PTC}                                                                                                                       & \multicolumn{5}{c}{NCI109}                                                                                                         \\
         & MSE $\downarrow$ & $\rho$ $\uparrow$ & $\tau$ $\uparrow$ & P@10 $\uparrow$  & P@20 $\uparrow$& MSE $\downarrow$ & $\rho$ $\uparrow$ & $\tau$ $\uparrow$ & P@10 $\uparrow$  & P@20 $\uparrow$& MSE $\downarrow$ & $\rho$ $\uparrow$ & $\tau$ $\uparrow$ & P@10 $\uparrow$  & P@20 $\uparrow$& MSE $\downarrow$ & $\rho$ $\uparrow$ & $\tau$ $\uparrow$ & P@10 $\uparrow$  & P@20 $\uparrow$& MSE $\downarrow$ & $\rho$ $\uparrow$ & $\tau$ $\uparrow$ & P@10 $\uparrow$  & P@20 $\uparrow$                        \\
         \cmidrule(r){2-6} \cmidrule(r){7-11} \cmidrule(r){12-16} \cmidrule(r){17-21} \cmidrule(r){22-26}
GMN      & 1.294 & 0.909 & 0.772 & 0.604 & 0.705 & 0.086 & 0.994 & 0.963 & 0.99 & 0.984 & 0.848 & 0.885 & 0.793 & 0.831 & 0.839 & 2.134 & 0.956 & 0.835 & 0.570 & 0.682 & 3.734 & 0.655 & 0.511 & 0.126 & 0.125  \\
SimGNN   & 2.057                       & 0.867                      & 0.716                      & 0.487                      & 0.588                      & 0.922                       & 0.976          & 0.907          & 0.942                      & 0.946                      & 0.481                       & \textbf{0.918} & \textbf{0.839}                      & 0.857                      & 0.874                      & 1.451                     & 0.948                      & 0.821                      & 0.559                      & 0.649                      & 3.650                       & 0.667          & 0.522                      & 0.104                      & 0.136                      \\
EGSC    & 1.595                       & 0.890                       & 0.746                      & 0.576                      & 0.655                      & 0.580                        & 0.984          & 0.954          & 0.969                      & 0.966                      & 0.431                       & 0.906          & 0.834                      & 0.880                       & 0.889                      & 6.964                     & 0.594                      & 0.454                      & 0.326                      & 0.436                      & 3.841                      & 0.669          & 0.525                      & 0.145                      & 0.174                      \\
ERIC     & 1.361                       & 0.904                      & 0.765                      & 0.629                      & 0.706                      & 0.147                       & 0.993          & 0.961          & 0.989                      & 0.990                       & 0.422                       & 0.891          & 0.820                       & 0.873                      & 0.873                      & 1.789                     & 0.905                      & 0.753                      & 0.532                      & 0.640                       & 3.833                      & 0.693 & 0.539                      & 0.150                       & 0.165                      \\
GraphSim & 1.581                       & 0.894                      & 0.751                      & 0.580                       & 0.675                      & 0.130                        & \textbf{0.994} & \textbf{0.966} & 0.987                      & 0.99                       & 0.515                       & 0.907          & 0.830                       & 0.863                      & 0.867                      & 1.883                     & 0.934                      & 0.797                      & 0.462                      & 0.620                       & 3.715                      & 0.678          & 0.533                      & 0.127                      & 0.158                      \\              
MGMN & 2.730 & 0.829 & 0.683 & 0.495 & 0.561 & 0.358 & 0.991 & 0.963 & 0.975 & 0.975 & 0.544 & 0.884 & 0.817 & 0.871 & 0.869 & 4.476 & 0.822 & 0.689 & 0.351 & 0.450 & 4.143 & 0.579 & 0.445 & 0.107 & 0.115 \\ 
\midrule
Ours & \textbf{1.037} & \textbf{0.925} & \textbf{0.795} & \textbf{0.701} & \textbf{0.751} & \textbf{0.044} & \textbf{0.994} & 0.960 & \textbf{0.994} & \textbf{0.997} & \textbf{0.299} & 0.904 & \textbf{0.839} & \textbf{0.901} & \textbf{0.903} & \textbf{0.993} & \textbf{0.962} & \textbf{0.850} & \textbf{0.601} & \textbf{0.717} & \textbf{3.568} & \textbf{0.701} & \textbf{0.549} & \textbf{0.177} & \textbf{0.195} \\
% Gain & \cellcolor[gray]{.85} 19.9\% & \cellcolor[gray]{.85} 1.8\% & \cellcolor[gray]{.85} 3.0\% & \cellcolor[gray]{.85} 11.4\% & \cellcolor[gray]{.85} 6.4\% & \cellcolor[gray]{.85} 48.8\% & \cellcolor[gray]{.85} 0.0\% & -0.6\% & \cellcolor[gray]{.85} 0.4\% & \cellcolor[gray]{.85} 0.7\% & \cellcolor[gray]{.85} 29.1\% & -1.5\% & \cellcolor[gray]{.85} 0.0\% & \cellcolor[gray]{.85} 2.4\% & \cellcolor[gray]{.85} 1.6\% & \cellcolor[gray]{.85} 31.6\% & \cellcolor[gray]{.85} 0.6\% & \cellcolor[gray]{.85} 1.8\% & \cellcolor[gray]{.85} 5.4\% & \cellcolor[gray]{.85} 5.1\% & \cellcolor[gray]{.85} 2.2\% & \cellcolor[gray]{.85} 1.2\% & \cellcolor[gray]{.85} 1.9\% & \cellcolor[gray]{.85} 18.0\% & \cellcolor[gray]{.85} 12.1\% \\
Gain & \textbf{19.9} \% & \textbf{ 1.8\% }& \textbf{ 3.0\% }& \textbf{ 11.4\% }& \textbf{ 6.4\% }& \textbf{ 48.8\% }& \textbf{ 0.0\% }&  -0.6\% & \textbf{ 0.4\% }& \textbf{ 0.7\% }& \textbf{ 29.1\% }&  -1.5\% & \textbf{ 0.0\% }& \textbf{ 2.4\% }& \textbf{ 1.6\% }& \textbf{ 31.6\% }& \textbf{ 0.6\% }& \textbf{ 1.8\% }& \textbf{ 5.4\% }& \textbf{ 5.1\% }& \textbf{ 2.2\% }& \textbf{ 1.2\% }& \textbf{ 1.9\% }& \textbf{ 18.0\% }& \textbf{ 12.1\%} \\
\bottomrule
\end{tabular}}
\caption{Experimental results of GSC. The bold represents the best performance. Gain is the percentage improvement of our model compared to the second place, and it is bolded to represent that our model improves performance on this metric. The experimental results show that our model achieves the best performance on 23 out of 25 metrics in five benchmark datasets compared to the state-of-the-art GSC method, which demonstrates the efficiency of our method.}
\label{exp:gsc}
\end{table*}

\subsection{Ablation Study}
\label{abl_study}
In this subsection, we validate the key design choices for our model by evaluating the impact of different components.

\textbf{Efficiency and Generalization of DiffAtt}. 
To verify the efficiency and generalization of DiffAtt, we evaluated it under four graph-level readout module settings. 
The four graph-level readout functions are Global Mean Pooling, Global Sum Pooling, Global Max Pooling and GCA. 
The temperature $t$ of DiffAtt here is set to 1 and the rest of the experimental setup and network architecture are the same as those used in Sec \ref{exp:gsl}. 
We report the MSE and $\rho$ metrics of each dataset in Table \ref{exp:diffatt_with_diff_pooling}. 
Surprisingly, \textbf{DiffAtt can effectively boost 37 of the 40 metrics on the five benchmark datasets for the four graph readout functions, which proves that DiffAtt has strong generalization abilities under different settings.} 
Specifically, on the MSE of the LINUX dataset, DiffAtt brings 43\% (0.541 vs. 0.311), 35\% (0.356 vs. 0.232), 49\% (0.122 vs. 0.062), and 58\% (0.115 vs. 0.048) improvement for each of the four graph readout settings, which demonstrates the efficiency of DiffAtt. 
GCA, due to its ability to dynamically adjust the weight of each node based on the global information, achieves better performance in most of the datasets. However, sum pooling works best specifically on NCI109.  \textit{Based on these results, we select to use GCA for our model architecture on all datasets except NCI109, where we use sum pooling.}

\textbf{The Impact of GNN Type, Enhancement of GIN, and Temperature Factor $t$}. Here, we evaluate the impact of using different types of GNNs, whether to use Residual Connections and FFN to enhance GIN layers, and the temperature factor $t$ of DiffAtt. The experimental results are shown in Table \ref{exp:gnn_enhance_learnable}. 
Compared with GCN and GAT, GIN has a better overall performance due to its theoretically stronger expressiveness. 
Residual Connections can help to improve the training process, while FFNs could enhance the embedding of nodes, thus augmenting GIN leads to a stable and huge enhancement effect. 
Compared to using fixed temperature factors, learnable temperature factors can give better results for most metrics. 
Based on these results, our final model uses a GIN encoder with Residual Connections and FFNs, along with a DiffAtt module with learnable $t$, which can reach the best performance in most cases.

\subsection{Main Results}
\label{exp:gsl}

We compare our \ourshort{} with a number of state-of-the-art GSC methods: GMN \cite{li2019graph}, SimGNN \cite{bai2019simgnn}, MGMN \cite{ling2021multilevel}, GraphSim \cite{bai2020learning}, EGSC \cite{qin2021slow} and ERIC \cite{zhuo2022efficient}. We reimplemented all baseline models based on the hyperparameters provided in the original paper. For hyperparameters that were not provided, several experiments were conducted to find the optimal parameter values. 

Table \ref{exp:gsc} presents the results of each model for a total of 25 metrics on the five benchmark datasets.
\textit{\textbf{Overall, our model achieves state-of-the-art performance on 23 of the 25 metrics.}} Especially on MSE, our model outperforms the second place by 19.9\%, 48.8\%, 29.1\%, 31.6\%, and 2.2\%, respectively. 
Compared to GMN, GraphSim and MGMN with complex node interaction modules, \ourshort{} achieves better performance with simple graph-level fusion modules, which illustrates the strong potential of graph-level embedding. Compared with other models using graph-level fusion modules, ERIC (NTN), SimGNN (NTN), and EGSC (EFN), \ourshort{} uses DiffAtt to enhance the difference information in the graph-level embedding, thus making the representation of the graph-level embedding more flexible and representative, and substantially improving the performance of the model. 
This proves the high performance of the \ourshort{}. 
Table \ref{exp:time_and_mem} compares the training time, GPU memory usage during training, and inference time of our graph-level fusion model \ourshort{} against other node-level fusion models on two datasets that have the largest number of nodes in the available datasets, PTC (25.6 nodes on average) and NCI109 (28.9 nodes on average). The node-level fusion models like GraphSim and MGMN have considerably higher training times and GPU memory consumption compared to \ourshort{}. In contrast, our \ourshort{} is more efficient, reducing training time by 3.7 times and using 6.2 times lower GPU memory across the datasets on average. Moreover, \ourshort{} has an advantage in inference speed. 
This efficiency advantage allows our method to scale to large-scale graphs.

\begin{table}[]

\centering
\resizebox{0.45\textwidth}{0.845cm}{
\begin{tabular}{ccccccc}
\toprule                
~ & \multicolumn{3}{c}{PTC}                 & \multicolumn{3}{c}{NCI109}                \\
                        & Train(s) & Mem(MiB) & Infer(s)& Train(s) & Mem(MiB) & Infer(s) \\
  \cmidrule(r){2-4}  \cmidrule(r){5-7} 
GMN                      & \textbf{259 $\pm$  1.41 }    & 1375   & 2.6 $\pm$   0.39         & 22646 $\pm$  332.34      & 1425   & 259.2   $\pm$ 1.43      \\
GraphSim                   & 1868 $\pm$ 9.19         & 9142   & 5.9  $\pm$   1.09       & 76231 $\pm$  348.60      & 9212   & 661.2    $\pm$  2.29    \\
MGMN                      & 1614  $\pm$ 1.41        & 16028  & 3.2    $\pm$  0.43      & 59858  $\pm$  485.08     & 13218  & 304.5    $\pm$  2.94    \\
\midrule
\ourshort{}                           & 299  $\pm$ 3.54        & \textbf{1235}   & \textbf{2.5 $\pm$ 0.43}            & \textbf{20210 $\pm$ 225.57}         & \textbf{1293}   & \textbf{235.1 $\pm$ 0.86}  \\
\bottomrule
\end{tabular}}
\caption{Experimental results of computational efficiency comparison between REDRAFT and node-level fusion models on larger graph datasets.}
\label{exp:time_and_mem}
\end{table}

\section{\evalshort{}: Quantifying Graph Fusion Modules' Ability}
\label{remaining_subgraph_test}

In this section, we propose a method called \textbf{\underline{Re}maining \underline{S}ubgraph \underline{A}lignment \underline{T}est (\evalshort{})} to quantify the ability of different graph fusion modules in capturing \unique{} information. 
Given graph pair $G_i$ and $G_j$ and a trained \ourshort{}, \textit{the core idea is to construct the \unique{} between $G_i$ and $G_j$ into a \textbf{single, valid graph} graph $\bar{G}_{i,j}$. Then, by testing the similarity between the fused embedding $F(E(G_i),E(G_j))$ and the \unique{} embedding $E(\bar{G}_{i,j})$, we can quantify how well the fusion module $F$ captures the \unique{} information to encode into the fused embedding. }

\begin{figure}[t]
  \centerline{\includegraphics[width=0.48\textwidth]{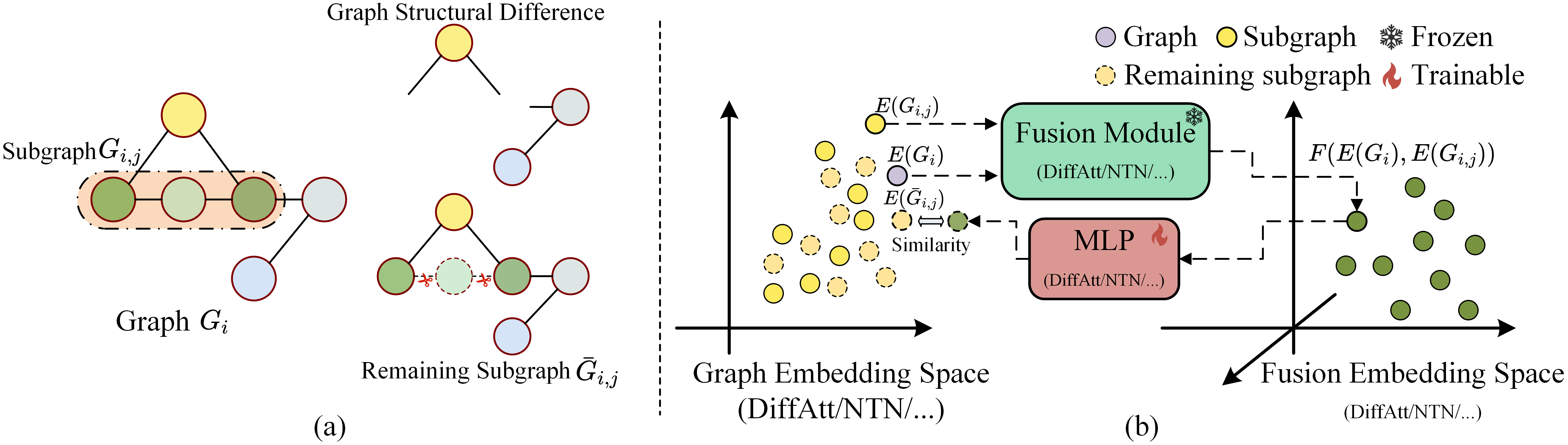}}
  \caption{A diagram of the \evalshort{}. (a) \evalshort{} constructs the \unique{} of $G_i$ and its subgraph $G_{i,j}$ into a remaining subgraph $\bar{G}_{i,j}$. (b) Based on the different trained GNN Encoder $E$ and fusion modules $F$, \evalshort{} leverages an MLP to test the maximum similarity between the fused embedding $F(E(G_i),E(G_{i,j}))$ and the \unique{} embedding $E(\bar{G}_{i,j})$.}
  \label{resat_pic}
\end{figure}

\textbf{Data Processing.} In order to construct \unique{} into a single valid graph structure, for each graph $G_i$ in the test set of the benchmark dataset, we construct subgraphs $G_{i,j}$ of it by sampling node subsets through random walks of length $\frac{N_i}{2}$ and including all edges between the sampled nodes. 
The proof provided in the appendix demonstrates that the mapping obtained when extracting subgraph $G_{i,j}$ from $G_i$ is the optimal mapping between them. 
As shown in Fig \ref{resat_pic}(a), the \textit{remaining subgraph} $\bar{G}_{i,j}$ is constructed by deleting all edges of $G_{i,j}$ in $G_i$ and then deleting all isolated nodes, which maximally contains the \unique{} between $G_i$ and $G_{i,j}$ in a single valid graph, and is therefore well suited to be used as a graph representing the \unique{} between $G_i$ and $G_{i,j}$. In total, we construct 50 such $(G_{i,j}, \bar{G}_{i,j})$ pairs per graph $G_i$. 

\textbf{Methodology.} We first train variants of REDRAFT on the benchmark datasets, replacing the fusion module $F$ with different graph-level modules like NTN, EFN, etc. After training, the parameters of each encoder $E$ and fusion module $F$ are fixed. 
However, the original fusion embeddings and \unique{} embeddings have different dimensions. Therefore, as shown in Fig \ref{resat_pic}(b), we train a new MLP regression model to map the embedding $F(E(G_i), E(G_{i,j}))$ to the target $E(\bar{G}_{i,j})$ to find their maximum similarity. For each fusion module $F$, the architecture of the MLP is adapted to optimize the mapping performance. 
The MSE loss between the fused graph embedding and the \unique{} embedding reflects their optimal similarity. A lower MSE suggests the fused embedding has already encoded abundant \unique{} features. This facilitates mapping it to the \unique{} embeddings, implying the fusion module effectively captures \unique{}. Conversely, a higher MSE indicates the fused embedding insufficiently represents \unique{}. Hence it is harder to align with the \unique{} embedding.

\begin{table}[t]

    \centering
    \resizebox{0.375\textwidth}{2.35cm}{
        \begin{tabular}{ccccccccccc}
        \toprule
        & \multicolumn{2}{c}{AIDS700nef} & \multicolumn{2}{c}{LINUX} & \multicolumn{2}{c}{IMDBMulti} \\
        ~ & GSC & \evalshort{} & GSC & \evalshort{} & GSC & \evalshort{}\\
        NTN & 3.131 & 2.009 & 0.602 & 1.989 & 0.436 & 2.611 \\ 
        No Fusion & 1.427 & 2.879 & 0.122 & 2.248 & 0.387 & 2.092 \\ 
        Square & 1.497 & 0.313 & 0.062 & 0.289 & 0.442 & 0.386 \\ 
        EFN & 1.346 & 1.024 & 0.070 & 0.950 & 0.392 & 0.901 \\ 
        Absolute & 1.344 & 0.204 & 0.070 & 0.155 & 0.365 & 0.750 \\ 
        \midrule
        Before DiffAtt & \multirow{2}*{\textbf{1.096}} & 1.008 & \multirow{2}*{\textbf{0.062}} &0.644 &\multirow{2}*{\textbf{0.316}} & 1.179\\
        After DiffAtt &~& \textbf{0.099} & ~&\textbf{0.105} & ~& \textbf{0.164}\\
        \midrule
        & \multicolumn{2}{c}{PTC} & \multicolumn{2}{c}{NCI109} & \multicolumn{2}{c}{\textit{ON AVERAGE}} \\
        ~ & GSC & \evalshort{} & GSC & \evalshort{} & GSC & \evalshort{} \\
        \cmidrule(r){2-3} \cmidrule(r){4-5} \cmidrule(r){6-7}
        NTN & 5.965 & 6.443 & 3.648 & 0.383 & 2.756 & 2.687 \\ 
        No Fusion & 1.539 & 2.057 & 3.650 & 3.276 & 1.425 & 2.510 \\ 
        Square & 1.936 & 5.562 & 3.478 & 0.569 & 1.483 & 1.424 \\ 
        EFN & 1.588 & 0.385 & 3.728 & 0.504 & 1.425 & 0.753 \\ 
        Absolute & 1.558 & 0.189 & 3.488 & 0.310 & 1.365 & 0.322 \\ 
        \midrule
                Before DiffAtt & \multirow{2}*{\textbf{1.040}} & 1.878          & \multirow{2}*{3.587} & 0.366          & \multirow{2}*{\textbf{1.220}} & 1.015 \\
        After DiffAtt  & ~                             & \textbf{0.181} & ~                    & \textbf{0.029} & ~                             & \textbf{0.116} \\
        \bottomrule
    \end{tabular} 
    }
\caption{Results of RESAT for various graph-level fusion modules, with the best result in bold. We also report the MSE accuracy of the model's graph similarity calculation (GSC). \textit{ON AVERAGE} denotes the average performance of each module on the five benchmark datasets. }
    \label{exp:fusion}
\end{table}

\textbf{Results.} The experimental results are shown in Table \ref{exp:fusion}. We also report results for \evalshort{} without any fusion module (No Fusion) and for graph-level fusion embedding before and after being adjusted by DiffAtt. Both NTN and No Fusion have poorer RESAT performance in most scenarios. However, NTN has lower accuracy in GSC because NTN focuses on the similarity of the two embeddings rather than the differences. Compared to No Fusion, EFN learns to capture \unique{} information during the learning process without strong guidance from the internal architecture. 
Absolute distance performs better than squared distance (on average 1.365 vs 1.483 on GSC; on average 0.322 vs 1.424 on \evalshort{}), probably because absolute distances better preserve the original distance information, whereas squared distances reduce the impact of smaller differences. It is found that the adjusted graph fusion embedding with DiffAtt achieved a substantial improvement of 88.6\% on average on \evalshort{} compared to before the adjustment (0.116 vs 1.015), suggesting that DiffAtt substantially enhance the information from \unique{}. 
We also find that the accuracy of GSC is closely correlated with the performance of \evalshort{}, with a $\rho$ of 0.899. This validates our hypothesis in Sec \ref{section:anaysis_on_graph_fusion} that better capturing \unique{} is key to modeling GED. 
DiffAtt outperforms the rest of the graph fusion modules on both GSC and \evalshort{}, confirming that DiffAtt can better access the information from \unique{} and thus better improve the performance of the model. 

% We also find that \evalshort{} performance is strongly correlated with GSC accuracy, confirming our design rationale of selectively enhancing \unique{}. 

\section{Conclusions and Limitations}
This paper presents a novel graph-level fusion module, \textbf{DiffAtt}, inspired by the observation that the value of GED is only related to the \unique{} between two graphs after optimal alignment. 
Based on DiffAtt, a new GSC model, \textbf{\ourshort{}}, is proposed to achieve state-of-the-art performance in the GSC task. 
DiffAtt uses the difference between the two embeddings to enhance the information of \unique{} in the graph-level embedding as attention, which substantially improves the performance of the model. 
Moreover, \ourshort{} uses Residual Connections and FFNs to augment powerful expressive GINs to generate better graph-level embeddings. 
Compared with other state-of-the-art GSC models, \ourshort{} achieves the best performance in 23 out of 25 metrics on five benchmark datasets. 
This paper also presents RESAT to quantify the ability of various graph-level fusion modules in capturing the \unique{}, and the results show that DiffAtt can better capture \unique{} information than other graph-level fusion methods. 

There are several limitations to our work, the first being that \ourshort{} as a GED predictor is not guaranteed to meet all properties of GED like triangular inequality and symmetry. Secondly, the additional use of FFNs in \ourshort{} reduces the speed of the model. Therefore, we encourage future work to address these issues and to better explore the potential of graph-level fusion.

\bibliography{aaai24}

\end{document}